\documentclass[conference]{IEEEtran}
\usepackage{csquotes}
\usepackage[numbers,sort&compress,square]{natbib}

\ifCLASSINFOpdf
  \usepackage[pdftex]{graphicx}
\else
\fi
\usepackage{amsfonts}
\usepackage[cmex10]{amsmath}
\usepackage{mathtools}

\newtheorem{definition}{Definition}

\usepackage{mdwmath}
\usepackage{mdwtab}

\makeatletter
\newif\if@restonecol
\makeatother

\usepackage[ruled,vlined]{algorithm2e}

\usepackage{fixltx2e}
\usepackage{url}
\usepackage{float}
\usepackage{placeins}

\newcommand{\parencite}{\citep}
\newcommand{\textcite}{\citep}
\newcommand{\AdaptAlgo}{\textsc{Adapt} }

\usepackage{hyperref}
\hypersetup{
    colorlinks = true,
    linkcolor = [rgb]{0.3,0.3,0.3},
    citecolor = [rgb]{0.3,0.3,0.3},
    urlcolor = [rgb]{0.3,0.3,0.3},
}

\begin{document}
%
\title{Diversity-Driven Selection of Exploration Strategies\\in Multi-Armed Bandits} 

\author{\IEEEauthorblockN{Fabien C. Y. Benureau}
\IEEEauthorblockA{Inria Bordeaux Sud-Ouest, FLOWERS Team\\
ENSTA ParisTech\\
Bordeaux University\\
fabien.benureau@gmail.com}
\and
\IEEEauthorblockN{Pierre-Yves Oudeyer}
\IEEEauthorblockA{Inria Bordeaux Sud-Ouest, FLOWERS Team\\
ENSTA ParisTech\\
pierre-yves.oudeyer@inria.fr}}

%

\newcommand\blindfootnote[1]{%
  \begingroup
  \renewcommand\thefootnote{}\footnote{#1}%
  \addtocounter{footnote}{-1}%
  \endgroup
}


\maketitle

\begin{abstract}
We consider a scenario where an agent has multiple available strategies to explore an unknown environment. For each new interaction with the environment, the agent must select which exploration strategy to use. We provide a new strategy-agnostic method that treat the situation as a Multi-Armed Bandits problem where the reward signal is the diversity of effects that each strategy produces. We test the method empirically on a simulated planar robotic arm, and establish that the method is both able discriminate between strategies of dissimilar quality, even when the differences are tenuous, and that the resulting performance is competitive with the best fixed mixture of strategies.
\end{abstract}
\blindfootnote{Originally published as: Benureau and Oudeyer, "Diversity-Driven Selection of Exploration Strategies in Multi-Armed Bandits," \emph{ICDL-EpiRob}, 2015. \href{https://doi.org/10.1109/devlrn.2015.7346130}{doi:10.1109/devlrn.2015.7346130}}


\section{Motivation}
We are given a black-box that takes inputs and produces outputs. We know the values the inputs can take, but we do not know which inputs produce which outputs. We do not even know which outputs \emph{can} be produced. We are given the opportunity to sample the black-box a limited number of times. In this context, we propose to investigate the following question: how much \emph{diversity of outputs} can be produced with the limited access we have?

This question defines an \emph{exploration problem}. Here, the objective is to discover what outputs the black-box is capable to deliver. To answer such a problem is to provide an \emph{exploration strategy}, i.e. a method that selects which inputs to experiment with on the black box, in order to produce a diversity of outputs.

In this paper, we interest ourselves with a scenario where we have multiple exploration strategies available, whose internal operational details are not specified, and we must select, for each available interaction with the environment---i.e., the black box---, which exploration strategy to use to generate the inputs to execute.

Stated differently, we have several exploration black-boxes and one environmental black-box, and we want to know which exploration black-box to use on the environmental black-box at each interaction, so as to maximize the diversity of the effects produced by the environmental black-box.

Two salient points are present in our problem statement. We consider \emph{exploration problems} rather than \emph{learning} ones. And we establish an objective of diversity, not one of control or of prediction or of fitness or of reward optimization. We briefly motivate these two stances in the following sections.

\subsection{Diversity and Exploration}

Behavioural diversity is a factor of individual robustness when facing an evolving environment. It ensures that the next time the environment changes some of the behaviours will remain relevant. At the population level, behavioural diversity provides variability even in the absence of genetic diversity.

This point was recently heeded by the evolutionary robotics community, which was facing, amongst others, two specific challenges: \emph{early convergence}, when the evolutionary process would get trapped in local minima because of a deceptive fitness function, and \emph{bootstrapping problems} where the first generation fails to produce rewarding behaviour, hence stalling the evolutionary process. The then solution, staging the fitness function \parencite{Gomez1997, Urzelai1998, Kodjabachian1998}---a method similar to \emph{reward shaping} in reinforcement learning \parencite{Dorigo1994, Mataric1994}---, was deemed impractical because requiring problem-specific fitness functions.

The solution came from replacing or modifying the fitness function to encourage \emph{behavioural diversity} in the population of candidate solutions \parencite{Trujillo2008, Lehman2008, Gomez2009, Delarboulas2010, Mouret2012, Doncieux2014}, a method proposed first in the classical evolutionary algorithm domain \parencite{Goldberg1987, Sareni1998}.

In infants, actively fostering diversity in our interaction with the environment through \emph{exploratory behaviour} is pivotal: it allows to discover and investigate new phenomena and affordances before they are detected as such. For Eleanor Gibson \parencite{Gibson1988}, babies are not endowed with the abilities to perceive affordances, but must spend their first years discovering affordances in their environment. For instance, children do not already know that mirrors are special objects proposing unique and salient interactions. Instead they must discover their affordance through an unrealed exploration of their environment. This point is important: studying exploratory behaviours on their own---rather than in the context of a learning problem---can shed light on how problems are discovered in the environment in the first place, \emph{before} they are acknowledged as learning activities.

One could argue that, after noticing the mirror particular nature, the exploratory behaviour of the child in front of the mirror is in fact highly structured, and follows the child-as-a-scientist paradigm \parencite{Schulz2007, Gweon2008, Gopnik2012}. But as Cook points out, more ecological explanations are also available: \enquote{selective exploration of confounded evidence is advantageous even if children explore randomly (with no understanding of how to isolate variables)} \parencite[p. 352]{Cook2011}. Therefore the mere production of behavioural diversity is a useful tool in broad and specific exploration.

One other reason to investigate exploration independently from learning: exploration can happen without learning. For instance, a robot randomly producing movements does not exhibit learning, yet exhibit an exploratory behaviour. Similarly, a robot following mindlessly the left wall of a maze explores the maze, and does it successfully to boot. And many vacuum robots available today explore their environment without learning them. In all those examples, exploration is present because the behaviour creates access to (new) information about the environment. That the information is not remembered or exploited is not an exploration issue, it is a learning one.

\subsection{More Than One Exploration Strategy}

Different environments lend themselves to different exploration strategies. In simple environments, doing random actions will be as effective as any other strategy. In more complex contexts, more elaborate strategies are needed.

The field of computational intrinsic motivation has developed an abundance of different motivational drives such as novelty, surprise, prediction error, predictive information or competence progress (see \parencite{Oudeyer2007, Baldassarre2013} for reviews). Each of these drives express preferences over \emph{what} is interesting in the world, and define specific exploration strategies.

Moreover, exploration, for a robot, may be possible through different means: asking for social guidance, observing a peer, or opting for self-exploration. Each of those venues may not be always available, and some, e.g. social guidance, may only be available for infrequent use.

This suggests that robots should be endowed with different exploration strategies to tackle complex environments. Furthermore, we argue that one should resist hiding the choice these strategies represent under a larger, monolithic, opaque exploration strategy. Indeed such a strategy would need to handle simultaneously \emph{how}, \emph{what} and possibly \emph{when} to explore, three aspects which may need to be specifically mediated by other components of behaviour.

Therefore, agents having multiple available exploration strategies are justified. In this article, we propose a strategy-agnostic method to select which strategy to choose in function of the empirical behaviour of each of them.

\vspace{1em}
\section{Problem}

\subsection{Environment}
\label{subsec:env}

An \emph{environment}, is formally defined as a function $f$ from $M$ to $S$. $M$ is the motor space, a bounded hyperrectangle of $\mathbb{R}^{m}$, and represents a parameterization of the movements the robot can execute. $S$ is the sensory space; it is a subset of $\mathbb{R}^{s}$. \emph{Effects} and \emph{goals}\footnote{We assume that $S$ is known by the exploration strategy, but nothing prevents $S$ to be set equal to $\mathbb{R}^{s}$} (desired effects) are elements of $S$.

A \emph{task} is defined as a pair $(f, n)$ with $f:M \mapsto S$ the environment and $n$ the maximum number of samples of $f$ allowed, i.e. the number of inputs the exploration strategy can try on the environment.

\subsection{Exploration}

An exploration strategy evaluates the function $f$, $n$ times, by providing a sequence of elements of $M$, $\mathbf{x}_0$, $\mathbf{x}_1$, ..., $\mathbf{x}_{n-1}$. Each $\mathbf{x}_i$ is evaluated as $\mathbf{y}_i = f(\mathbf{x}_i)$, and $\mathbf{y}_i$ is observed by the exploration strategy before $\mathbf{x}_{i+1}$ is chosen.

In order to evaluate the exploration strategy, we use an \emph{exploration measure} $\mathcal{C}$, that takes the behavioural trace of the agent as input, i.e., the actions executed and effects produced: $\{(\mathbf{x}_i, \mathbf{y}_i)\}_{0 \leq i < n}$.

A common objective of the experimenter is to evaluate if the agent has obtained knowledge of all the possibilities of the environment. A good proxy for this is to evaluate the set of effects the agent was able to produce during the exploration. In other words, how well the image of $f$, $f(M)$---the \emph{reachable space}---was sampled.

Since we do not assume that the agents have knowledge of the exploration problem they are examined under, or that they have knowledge of the exploration measures that are used to evaluate their behaviour, and since agents may explore the environment for their own purposes, and self-evaluate their behaviour according to their own metrics, the choice of an exploration measure is necessarily arbitrary. This consideration is not present for instance in reinforcement learning, where the cumulative reward defines an objective motivation for the agent, and an objective evaluation for the experimenter. In an exploration context, it is the responsibility of the experimenter to justify the interest and relevance of the selected exploration measure.

In this work, we select a \emph{diversity measure} to evaluate the exploration. The importance of diversity for the development of humans and animals was argued above. And behavioural diversity has proven itself empirically in the field of evolutionary robotics. Absent an objective environmental reward for the agent's behaviour, and absent an assumption that the agent possesses specific learning abilities, encouraging diversity in behaviour is relevant in multiple ways. First, it does not put tight constraints on the structure of the behaviour of the agent. Second, it prepares the agent for future problems: an agent with a diverse behavioural repertoire is likely to also have high amounts of diverse knowledge and skills.

The diversity measure concerns itself only with the sensory part of the behaviour: $\{\mathbf{y}_i\}_{0 \leq i < n}$. It is defined as a coverage measure. Given $\tau > 0$, the diversity of the exploration $\mathcal{C}(\{\mathbf{y}_i\}_{0 \leq i < n})$ is defined as the volume (more precisely the Lebesgue measure) of the union of the $n$ hyperballs of $\mathbb{R}^{s}$ with $\mathbf{y}_0$, $\mathbf{y}_1$, ..., $\mathbf{y}_{n-1}$ as a centres, and radius $\tau$.
\[
\mathcal{C}_{\tau}(\{\mathbf{y}_i\}_{0 \leq i < n} = \bigcup_{i=0}^n{B(\mathbf{y}_i, \tau)}
\]with $B(\mathbf{y}_i, \tau)$ the hyperball of radius $\tau$ and centre $\mathbf{y}_i$.

In evolutionary robotics, other measures of diversity such as sparseness \parencite{Lehman2008} or entropy \parencite{Delarboulas2010} have been used.

\vspace{1em}
\section{Illustrating the Problem}
In this section, we illustrate the problem on a specific example, that will serve as the experimental setup for the method, exposed in the next section.

We consider an idealized robotic arm on a two-dimensional plane, made up of an open chain of 20-joints linked by segments of 1/20th of a meter each, so that the total length of the arm is one meter. The angles of the joints are restricted to values between -150 and 150 degrees. The angles of the joints are the inputs: they uniquely define the posture of the arm, and therefore, the position of the end-effector, which corresponds to the environmental feedback. Let's remark that only the final position of the end-effector, corresponding to the angle inputed in absolute value, is returned by the environment (i.e. there is no posture dependence between two consecutive samples).

\begin{figure}[!b]
\centering
\includegraphics[width=0.33\textwidth]{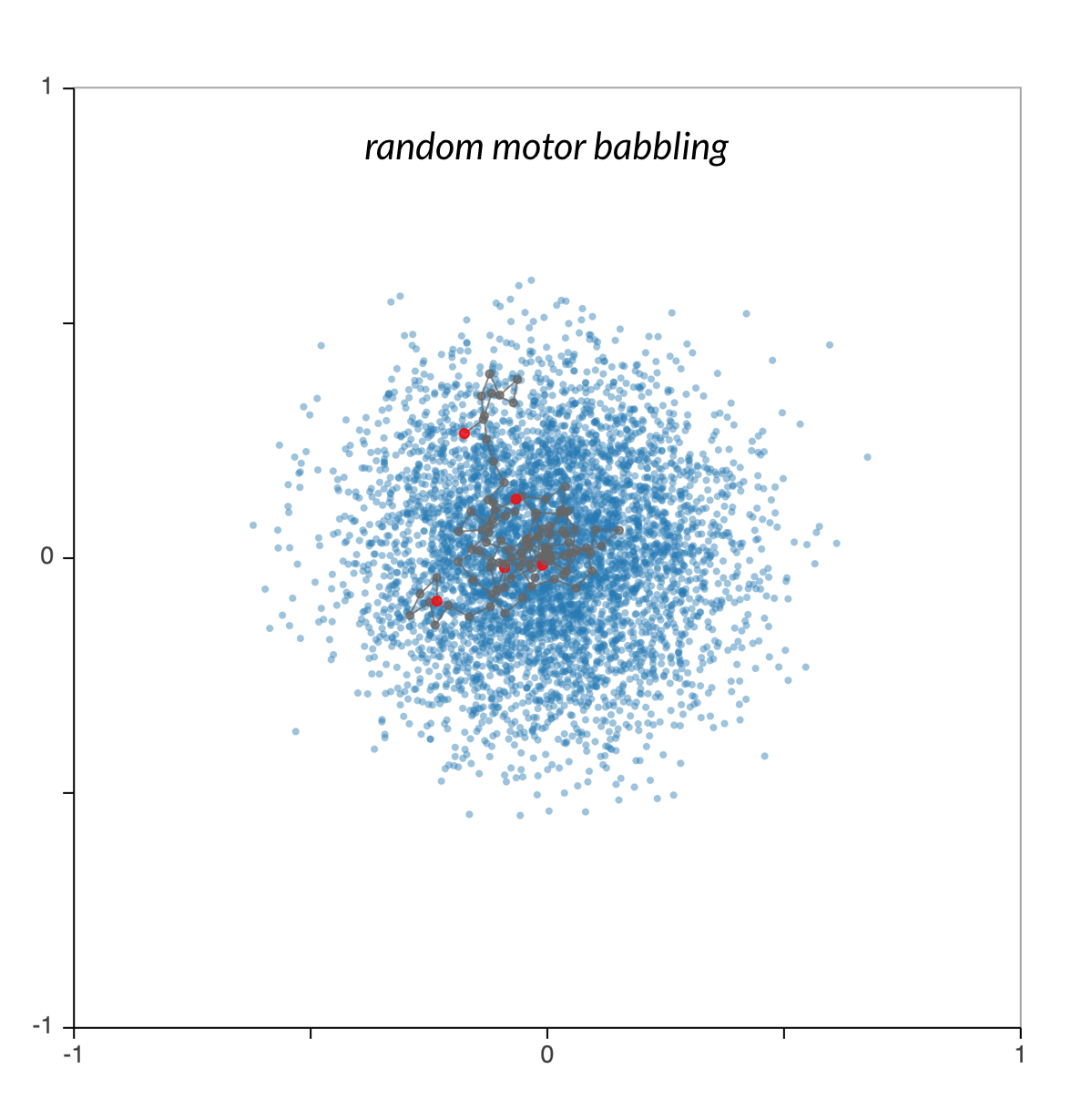}
\caption[Random Motor Babbling]{Random motor babbling is not an efficient exploration strategy with a high number of joints.\label{fig:kin_mb}}
\end{figure}

\begin{figure}[!b]
\centering
\includegraphics[height=0.79\textheight]{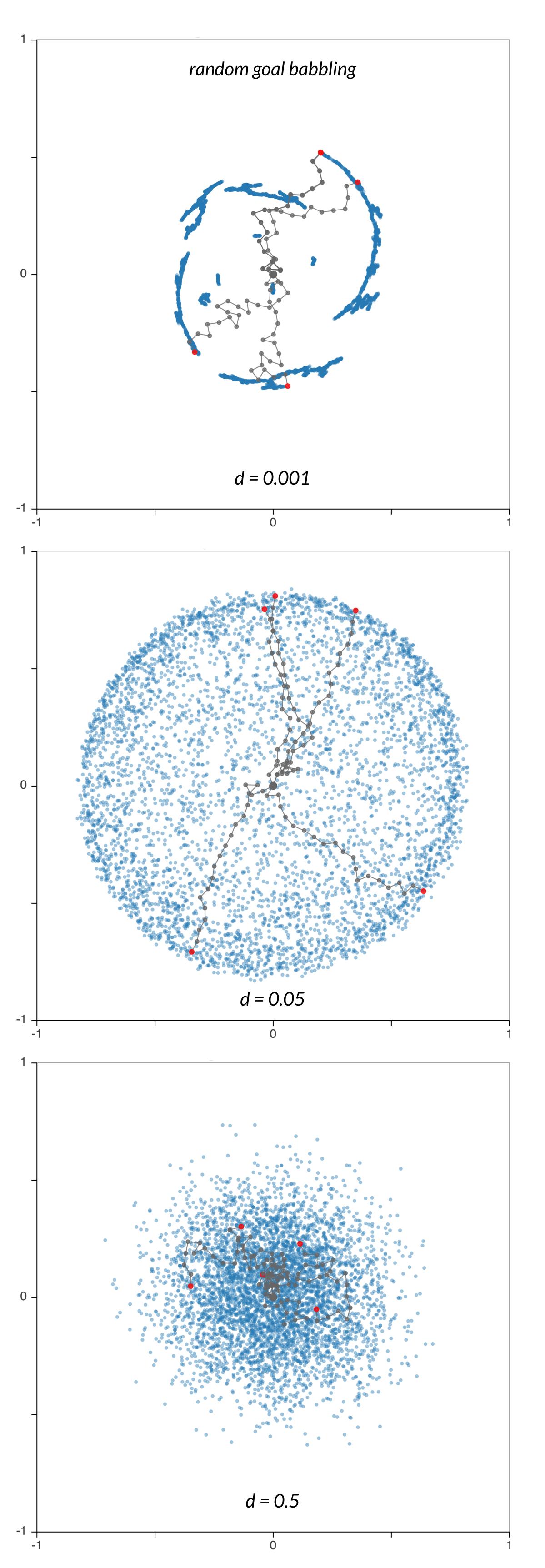}
\caption[Random Goal Babbling]{Random goal babbling can be a very efficient strategy---if the inverse model is well chosen. Each exploration is done over 5000 timesteps. In each case, the last five postures of the exploration are displayed.\label{fig:kin_gb}}
\end{figure}

\subsection{A Tale of Two Exploration Strategies}

Despite the simplicity of the arm setup, it is not a trivial problem, and this is exacerbated since we cannot assume any knowledge about the arm.

The most simple strategy, \emph{random motor babbling} (RMB), samples the motor space randomly. Here the RMB strategy (Figure \ref{fig:kin_mb}) is inefficient: indeed, the redundancy\footnote{Considering a subset of the sensory space $B$, the \emph{redundancy} of $B$ is defined as the volume (more generally, the Lebesgue measure) of the set of motor commands whose effect belong to $B$, i.e. $\{\mathbf{x}| f(\mathbf{x}) \in B \}$ with $f$ the environment feedback function (see section \ref{subsec:env})). \textcite{Lenarcic1999} provides an algorithm to quantify the redundancy of rigid, multijoint robotic arms, but the computation is only tractable for a small number of joints.} of the arm is heterogeneously distributed in the sensory space (the end-effector position space). In particular, the redundancy is high near the origin, and order of magnitude lower on the edge of the reachable space. Because the RMB strategy is precisely an estimator of the heterogenity of the redundancy, it rarely ever explores the edges of the reachable space.

A goal babbling strategy is (usually) better suited for exploring the arm setup. We will consider a \emph{random goal babbling} (RGB) strategy \parencite{Rolf2011, Baranes2010}, that picks a goal at random in the square $[-1, 1]\times[-1, 1]$, and translate it to a tentative motor command that tries to put the end-effector as close as possible of the goal.

To translate a goal into a motor command, we need an \emph{inverse model}. In this paper, we are only interested in relative performance: we choose a simple inverse model. Our inverse model, when given a goal, finds the nearest effect available in the observed data, retrieves the motor command that produced it, applies a small perturbation to it, and returns the perturbed command for execution of the exploration strategy. The magnitude of the perturbation is parametrized by the \emph{perturbation coefficient} $d$: the perturbation is randomly chosen between $\pm d$ times the legal joint range (here $300^\circ$). For instance, if $d=0. 05$, the motor command is perturbed by a random value chosen in $\pm 15^\circ$ on each joint.

Choosing $d$ appropriately is not trivial. In Figure \ref{fig:kin_gb}, the RGB exploration of three different values of $d$ is shown. The $d=0.05$ case results in a good exploration. But $d=0.001$ creates degenerated clusters: the perturbation is too low to create enough sensory variability. \emph{A contrario}, $d=0.5$ creates too much variability, and is only marginally better than the RMB exploration of Figure~\ref{fig:kin_mb}.

Let's imagine now that we are given two strategies to explore the arm setup. One is the RMB strategy, and the other is a RGB strategy, with unknown $d$. We don't assume any knowledge of either strategy. How can we dynamically decide, for each interaction with the black-box, which exploration strategy to choose to maximize the coverage of the exploration over the reachable space?

\FloatBarrier

\subsection{Inverse Model}
\label{subsec:inverse_model}

Given a goal, the inverse model finds the nearest neighbour in the observed effects and applies a small perturbation on its corresponding motor command.

Formally, $M$ is an closed hyperrectangle of $R^m$, and as such it is the Cartesian product of $m$ closed intervals:
\[M = \prod_{m=0}^{m-1} [a_i, b_i]\]
Given an motor command $\mathbf{x} = \{x_0, x_1, ..., x_{m-1}\}$ in $M$, a perturbation of $\mathbf{x}$ is defined by:
\begin{align*}%
\textsc{Perturb}_d(\mathbf{x}) = \{\textrm{random}(&\textrm{max}(a_j, x_j-d(b_j-a_j)),\\
                                                   &\textrm{min}(x_j+d(b_j-a_j), b_j))\}_{0 \leq j < m}
\end{align*}%
with the function \emph{$\textrm{random}(a, b)$} drawing a random value in the interval $[a, b]$ according to a uniform distribution. $d$ is the \emph{perturbation parameter}, and the only parameter of the inverse model, that we can now express in Algorithm \ref{algo:inverse}.

\begin{algorithm}
\DontPrintSemicolon
\KwData{
\begin{itemize}
\item $d \in [0, 1]$, a perturbation ratio.
\item $E = \{(\mathbf{x}_t, \mathbf{y}_t)\}_{0 \leq t < N} \in {(M\times S)}^N$, past observations.
\item $\mathbf{y}_g \in S$, a goal.
\end{itemize}
}

\KwResult{
\begin{itemize}
\item $\mathbf{x}' \in M$ a motor command.
\end{itemize}}
\;
Find $(\mathbf{x}_i, \mathbf{y}_i)$ in $E$ so that $\mathbf{y}_i$ is the the nearest neighbour of $\mathbf{y}_g$ in $\{\mathbf{y}_t\}_{0 \leq t < N}.$\;
$\mathbf{x}' = \textsc{Perturb}_d(\mathbf{x}_i)$
\caption{$\textsc{Inverse}_d(\mathbf{y}_g, E)$\label{algo:inverse}}
\end{algorithm}

\vspace{1em}
\section{Method}

\subsection{Effect Diversity}

Choosing which strategy to employ at each step of the exploration faces three main challenges:
\begin{enumerate}
\item \textbf{Interdependence}: an exploration strategy effectiveness may depend on another strategy; goal babbling relies on motor babbling to bootstrap the exploration. Given the inverse model currently used, this is even more true, as goal babbling's performance depends heavily the sensorimotor attractors in which it expands, and thus on the location of the observations produced early in exploration by motor babbling.
\item \textbf{Dynamical Value}: the usefulness of a strategy may change rapidly. Motor babbling is useful in the beginning of the exploration, but its usefulness drops quickly.
\item \textbf{Agnosticity}: since an exploration strategy might be arbitrarily complex, and possibly involve, in turn, other exploration strategy, an adaptive strategy should not rely on knowledge of the internal workings of the strategies amongst which it must choose.
\end{enumerate}

Interdependence does not have to be handled directly, but it implies that even strategies that did poorly in the past must be re-evaluated regularly as the exploration progresses. The dynamical nature of the contribution of each strategy means that performance data becomes obsolete quickly: evaluations should be done over short-term time windows. Agnosticity implies the contributions of the strategies have to be evaluated only from the observations the strategies produce. We introduce a measure that matches those constraints now.

A strategy that produces effects over areas that have already been explored is of little use for exploration. We introduce an online \emph{diversity measure} that evaluates, each time a strategy is used, how much diversity is created, with regards to already observed effects.

In order to do that, we rely on the diversity measure introduced in section II, based on the union of disks centred on observed effects. Although we reuse the coverage measure here out of convenience, the two measures do not have to have any relationship with one another. The measure is adapted to evaluate a single effect: the diversity of a new observed effect is the increase in diversity, i.e., the increase in the covered area.

\begin{definition}
Given a set of effects $E=\{\mathbf{y}_0, \mathbf{y}_1, ..., \mathbf{y}_{n-1}\}$, and a coverage threshold $\tau$ in $\mathbb{R}^+$, the diversity of a new effect $\mathbf{y}_{n}$ relative to $E$ is defined as:
\[\textrm{div}_{\tau}(\mathbf{y}_n, E) = \mathcal{C}_{\tau}(E \cup \{\mathbf{y}_n\})-\mathcal{C}_{\tau}(E)
\]
\end{definition}

The diversity of a strategy, in turn, is the averaged diversity of the effects it produced, over a given time window.

\begin{definition}
Given a set of strategies $s_0, s_1, ..., s_{q-1}$, and a set of observed effects $E = \{\mathbf{y}_0, \mathbf{y}_1, ..., \mathbf{y}_n\}$, we have for a given strategy $s_j$ a subsequence $\mathbf{y}^j_0, \mathbf{y}^j_1, ..., \mathbf{y}^j_{n_j}$ of the effects produced by motor commands emanating from the strategy. Given a time window $w$ in $\mathbb{N}^+$, we define the diversity of strategy $s_j$ as:
\[
 \text{div}_{\tau, w}(s_j, E) =
  \begin{dcases}
   \frac{1}{w'}\sum_{i=0}^{w'}{\textrm{div}_{\tau}(\mathbf{y}^j_{n_j-i}, E)} &\text{if } n_j > 0 \\
   0          & \text{otherwise}
  \end{dcases}
\]
with $w'=\textrm{min}(w, n_j)$.
\end{definition}

\subsection{Multi-Armed Bandits}

As expressed above, the problem we tackle shares similarities with the Multi-Armed Bandit problem (MAB) \parencite{Robbins1952}. The exploration strategies are the bandits, amongst which the agent must choose to create diversity. However, the feedback received is a sensory feedback from the environment, which cannot be used as is in the MAB setting.

Using the diversity measure of a strategy introduced above, we can now evaluate the contribution of each strategy to the exploration. We now have a classic MAB problem: we choose between a finite number of different strategies with different diversity scores, and after choosing one we receive a feedback signal from the chosen strategy from which we compute an updated score.

The classic MAB problem considers only bandits that are independent from one another (choosing one does not affect the value of the others), and stationary (the distribution of rewards of the bandit does not change). A variation of the problem, the \emph{adversarial} (also called \emph{non-stochastic} or \emph{non-stationary}) MAB, removes the stationary and interdependence assumptions: an adversary is free to choose arbitrary rewards for each bandit at each timestep.

In practice, a significant portion of the published literature on the adversarial MAB problem only removes the stationary assumption. In other words, the problem takes place in the \emph{oblivious} opponent model: the actions of the adversary, i.e. the rewards for each bandit at each timestep, are decided before the game starts. This is the case in \textcite{Whittle1988} and \textcite{Auer2002}, who investigate rewards that can arbitrarily change. \textcite{Garivier2008} presents \emph{abruptly changing environments}, where all bandits' reward distributions change at specified timesteps. \textcite[156--169]{CesaBianchi2006} provides a treatment of the nonoblivious case.

Recently, \textcite{Lopes2012} introduced the \emph{Strategic Student Problem} that tries to capture the issues involved when learning multiple tasks at the same time. A student has to learn multiple topics (maths, chemistry, history, etc.), and has limited resources (time) to do so. How should he allocate his study time between topics in order to maximize its mean grade at the end of the semester? A possibility is to consider the problem as a MAB problem where the bandits are learning tasks. Interestingly, the works of \textcite{Baranes2010} on goal babbling can be understood in this perspective: each region of the goal space is a different topic, whose improvement is empirically measured through competence progress during learning, and the exploration strategy must decide how to distribute its action given those learning feedback signals.

The strategic student problem also considers another related problem: a student has one topic to learn, but several possible learning strategies. Which one should he choose? Is a mixture of several strategies better than employing the best one all the time? This is the problem of learning \emph{how} to learn \parencite{Schmidhuber1995}. \textcite{Baram2004} explored such a problem and showed that a dynamically selected mixture of three active learning strategies outperformed any pure strategy. \textcite{Konidaris2008} demonstrated that empirically evaluating and selecting among different small state space representations specific to a task during learning was effective and avoided a large task space when learning was unfeasible. The work of \textcite{Nguyen2012} investigates robots dynamically choosing between asking a teacher for a demonstration or doing self-exploration on their own. \textcite{Jauffret2013} proposes a method where a robot can self-assess, and has a frustration drive. When frustrated, the robot can opt to choose social help to improve its performance. In the context of reinforcement learning, \textcite{Hester2013} develops an algorithm that can evaluate dynamically which exploration strategy brings the most rewards. These exploration strategies are driven by extrinsic and intrinsic motivations: maximizing rewards, reducing variance, seeking novelty, seeking unexplored states (a binary novelty), and seeking or avoiding particular features of the state representation. \textcite{Clement2015} uses the framework of the Strategic Student Problem to create a tutoring system that actively personalizes the sequence of activities to each student, by tracking their performance and identifying which exercises and modalities make the student progress faster. The works of \parencite{Baram2004}, \parencite{Nguyen2012} and \parencite{Hester2013} are singular because they combine deciding \emph{how} to learn, and deciding \emph{what} to learn, using a hierarchical approach. The learning strategy is selected first (\emph{how}), and then it chooses what input to sample (\emph{what}).

Learning performance typically exhibits \emph{diminishing returns}, and \textcite{Lopes2012} shows that, in the strict case, this allows to express the mean performance across tasks as a \emph{submodular function} \parencite{Krause2014}.  \parencite{Nemhauser1978} has proven that with non-decreasing submodular function, the greedy strategy is guaranteed to be no worse than $1-\frac{1}{e} \approx 0.63$ times the optimal solution for maximizing the function. Of course, not all set of learning tasks exhibit a submodular structure. Still, it suggests that a good-enough performance might be obtained through simple-enough algorithm in practice. \textcite{Lopes2012} and \textcite{Hester2013} advocate the use of the EXP4 algorithm \parencite{Auer2002} rather than a greedy algorithm, as a more robust approach.

Compared to these works, our approach distinguishes itself on two fronts: first, we are selecting exploration strategies to improve exploration, rather than exploration or learning strategies to improve learning. The resulting strategy is another exploration strategy. Second, we are using diversity to transform the feature vector of the sensory feedback into a scalar that can be adequately interpreted as a reward. To our knowledge, this is the first work to do that in the context of a Multi-Armed Bandit problem.

\subsection{Adaptive Strategy}

\begin{algorithm}[b!]
\DontPrintSemicolon
\SetKwInOut{Input}{Input}

\Input{
\begin{itemize}
\item $s_0, s_1, ..., s_{q-1}$, strategies.
\item $E=\{\mathbf{y}_0, \mathbf{y}_1, ..., \mathbf{y}_{n-1}\}$, a set of effects.
\item $\tau$, coverage threshold.
\item $w$, time window.
\item $\alpha$, ratio of random choice.
\end{itemize}
}

\KwResult{
\begin{itemize}
\setlength\itemsep{0em}
\item $s_j$, chosen strategy
\end{itemize}
}
~

\eIf{{\sc{Random()}} $< \alpha$}{
    choose a random strategy.\;
}
{
    choose a strategy $s_j$ proportionally to its diversity $\text{\emph{diversity}}_{\tau, w}(s_j, E)$.\;
}

\caption{{\sc{Adapt}}($w$, $\tau$)\label{algo:Adapt}}
\end{algorithm}

The \AdaptAlgo algorithm chooses strategies proportionally to their diversity. To allow for constant re-evaluation of the strategies, even those with low diversity, the algorithm chooses a strategy  at random $\alpha$ percent of the time, with $\alpha > 0$. Algorithm \ref{algo:Adapt} formally describes this.

Additionally, in order to foster initial experimentation with each strategy, the diversity measure is overestimated at the beginning of the exploration. For a given strategy $s_j$, instead of considering the set $E_j = \{\mathbf{y}_0, \mathbf{y}_1, ..., \mathbf{y}_{n_j}\}$, we consider the set $E' = \{\mathbf{y}_{-k}, \mathbf{y}_{-k+1}, ..., \mathbf{y}_0, ..., \mathbf{y}_n\}$, with $k$ in $\mathbb{N}^+$. The set $\{\mathbf{y}_{-k}, \mathbf{y}_{-k+1}, ..., \mathbf{y}_{-1}\}$ is composed of fictitious points only available to the selecting strategy, that generate hyperballs that do not overlap with the observed effects. That way, the diversity of the strategy is overestimated during the $w$ first times it is selected. This also avoids having the first strategy selected unfairly preferred because it created the first observation, thus receiving the diversity of a full hyperball volume. We will use $k = 1$ in all strategies.

\section{Results}
\begin{figure*}[!p]
\centering
\includegraphics[width=\textwidth]{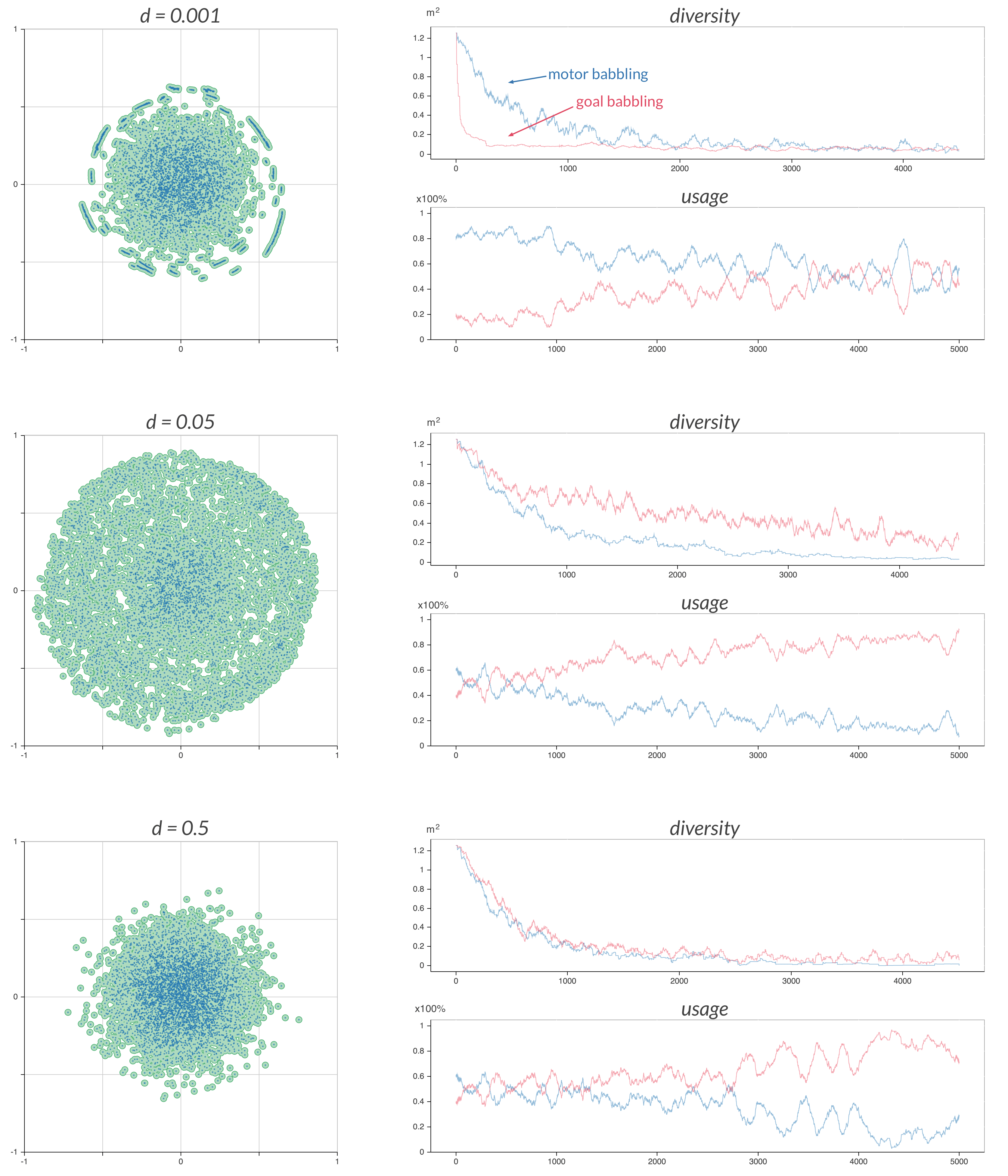}
\caption[Qualitative Results]{The \AdaptAlgo algorithm correctly selects the best strategy in all three contexts. For each learner, three graphs are shown: the spread graph with the coverage area ($\tau = 0.02$), the diversity graph giving the diversity measure of each strategy in function of the timesteps, and the usage graph, showing how the strategies are effectively used. For the usage graph, the data at time $t$ shows the percentage of use averaged over the surrounding 100 timesteps (50 before, 50 after).\label{fig:adapt_qual}}
\end{figure*}

\begin{figure}[!h]
\centering
\includegraphics[width=0.5\textwidth]{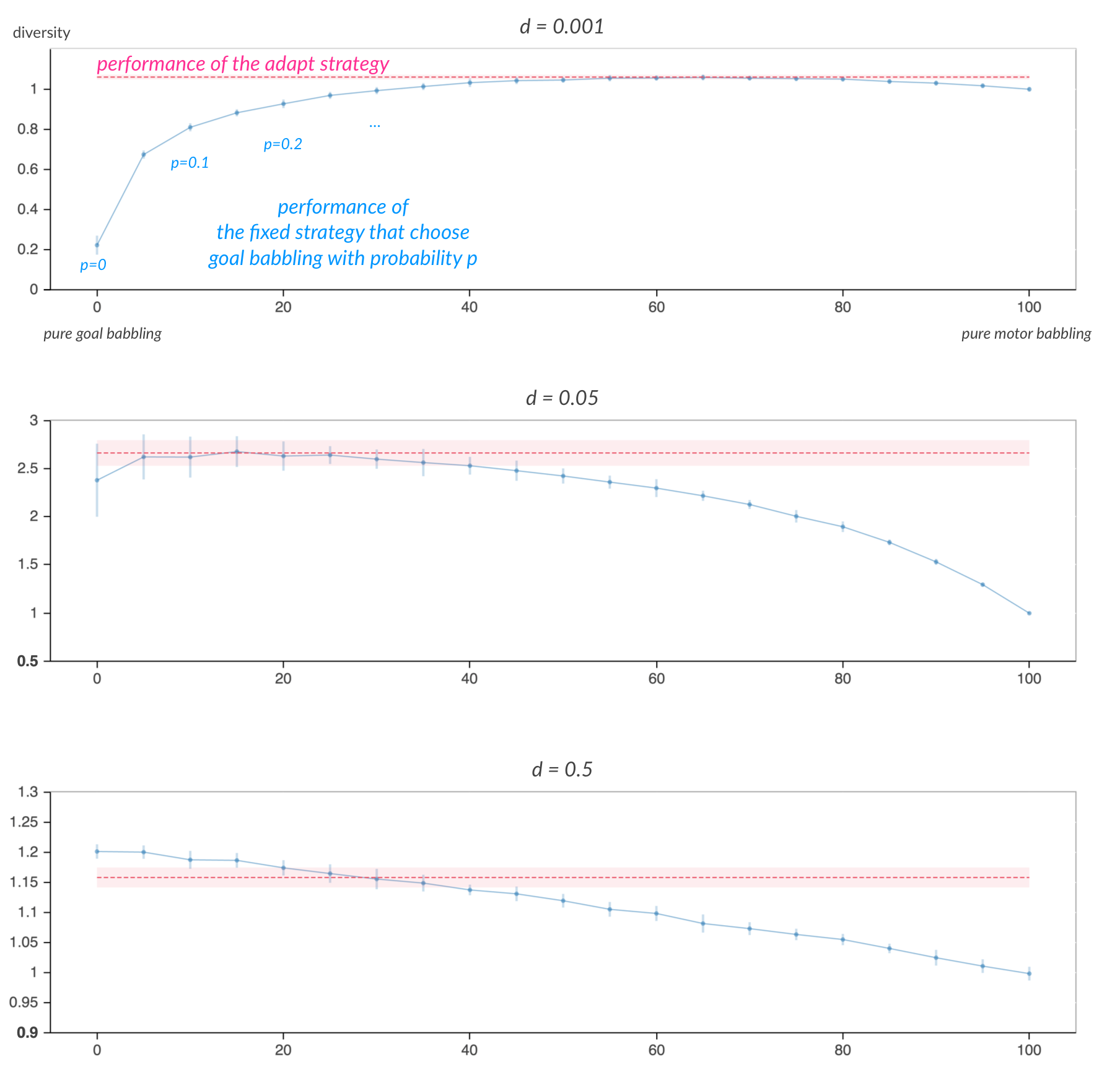}
\caption[Quantitative Results]{The \AdaptAlgo algorithm performs well when strategies behave distinctly, and better than random with similar strategies. Each graph displays the performances of fixed mixtures of the two strategies, with the performance of the adaptive strategy added as a dotted line (its standard deviations in displayed in light colour as well). Experiments were repeated 25 times. Note that not all the y-axis of the graphs begin at zero.\label{fig:adapt_quant}}
\end{figure}

Figure~\ref{fig:adapt_qual}, the results of the strategy are displayed. In all three learner configurations, the \AdaptAlgo algorithm identifies and uses the correct strategies. When $d = 0.001$, the goal babbling strategy is inefficient in the beginning, and motor babbling is overwhelmingly used. Motor babbling diversity declines continually during the exploration, and in the later stage, is comparable to goal babbling. As a result, after 4000 timesteps, the two strategies are used roughly equally.

When $d = 0.05$, goal babbling and motor babbling produce the same diversity at the beginning, but goal babbling declines more slowly than motor babbling. As a result, goal babbling is used more and more as the exploration progresses, as it should be.

When $d = 0.5$, motor and goal babbling behave similarly---if $d$ had been equal to $1.0$, they would be the same strategy. During the early phase of the exploration, the \AdaptAlgo algorithm does not distinguish between the two strategies. But in the later phase, goal babbling is able to provide an edge, however small, that is detectable by the \AdaptAlgo algorithm. Goal babbling usage dominates after 1500 timesteps, and is used 80\% of the time after 4000 timesteps.

While the algorithm works qualitatively, it remains to be seen if this translates quantitatively. Figure \ref{fig:adapt_quant} compares the error of the \AdaptAlgo algorithms with fixed-ratio strategies, where the motor babbling strategies is chosen with probability $p$, and the goal babbling with probability $1-p$.

When goal babbling is much worse than motor babbling ($d = 0.001$) or when it is much better ($d = 0.05$), the \AdaptAlgo algorithm manages performance on par with the best fixed mixture of strategies. When goal and motor strategy behave similarly, the adapt strategy is more conservative than the best case. This stems from the early stage of the exploration, when the motor babbling and goal babbling strategies are both effective, and hence both significantly used.

\section{Discussion}
The \AdaptAlgo algorithm we proposed, and the corresponding adaptive strategy we implemented demonstrate how a choice of multiple exploration strategies can be exploited to explore an unknown environment. The diversity measure is, in many ways, rather crude, but it shows that discriminating between exploration strategies is definitely possible, and, advantageous. The general idea behind this work is not particularly new.

Its application to a diversity measure is, however. In fact, since exploration, as explained, does not make the typical assumption about the agents capabilities---it does not assume the agent is capable (or willing) to make predictions, nor to exert (or demonstrate) control over the environment---the method we presented extends the applicability of the Multi-Armed Bandits to situations where learning or reward signals are not present. And it does so without requiring to design a problem-specific reward function.

Our work could be criticized for the simplicity of the environment that is used, and that's a valid point. Yet, we chose to present this method on a simple setup here to avoid the reader having to suspend his intuition, or suspect interference from the robot complex dynamics into the results. The extreme simplicity of our inverse model is also a deliberate choice in this regard. We are currently preparing experiments on a real robot actuated through dynamical motor primitives to reproduce the results in a more complex scenario.

From the experiments we conducted, it is unclear how the \AdaptAlgo algorithm will scale with the number of strategies. As more strategies are available, either more time will have to be devoted to exploratory sampling of bad strategies, or strategies will be less accurately evaluated overall. This is the classic exploration/exploitation trade-off.

\section*{Acknowledgments}

This work was partially funded by the ANR MACSi and the ERC Starting Grant EXPLORERS 240 007. Computing hours for running simulations were graciously provided by the MCIA Avakas cluster.

\section*{Source Code}

The complete source code behind the experiments and the figures is open source and available at:\\ \href{https://doi.org/10.6084/m9.figshare.6809135}{doi.org/10.6084/m9.figshare.6809135}.

\def\bibfont{\footnotesize}
\def\bibfont{\fontsize{7pt}{7pt}\selectfont}
\bibliographystyle{plain}
\bibliography{references}

\end{document}